\documentclass{article}

\usepackage{microtype}
\usepackage{graphicx}
\usepackage{subfigure}
\usepackage{booktabs} 

\usepackage{hyperref}

\usepackage{colortbl}
\usepackage[dvipsnames]{xcolor}

\usepackage[accepted]{icml2023}

\usepackage{amsmath}
\usepackage{amssymb}
\usepackage{mathtools}
\usepackage{amsthm}
\usepackage{graphicx}

\usepackage[capitalize,noabbrev]{cleveref}

\theoremstyle{plain}

\theoremstyle{definition}

\theoremstyle{remark}

\usepackage[textsize=tiny]{todonotes}

\icmltitlerunning{No Free Lunch in Self Supervised Representation Learning}

\begin{document}

\twocolumn[
\icmltitle{No Free Lunch in Self Supervised Representation Learning}



\icmlsetsymbol{equal}{*}

\begin{icmlauthorlist}
\icmlauthor{Ihab Bendidi}{ens1,minos}
\icmlauthor{Adrien Bardes}{ens2,fb}
\icmlauthor{Ethan Cohen}{ens1,syn}
\icmlauthor{Alexis Lamiable}{ens1}
\icmlauthor{Guillaume Bollot}{syn}
\icmlauthor{Auguste Genovesio}{ens1}
\end{icmlauthorlist}

\icmlaffiliation{ens1}{IBENS, Ecole Normale Supérieure, PSL Research Institute, Paris, France}
\icmlaffiliation{ens2}{INRIA, Ecole Normale Supérieure, PSL Research Institute, Paris, France}
\icmlaffiliation{minos}{Minos Biosciences, Paris, France}
\icmlaffiliation{fb}{FAIR, Meta, Paris, France}
\icmlaffiliation{syn}{Synsight, Evry, France}

\icmlcorrespondingauthor{Auguste Genovesio}{auguste.genovesio@ens.psl.edu}

\icmlkeywords{Self Supervised learning, Representation learning, Transformations optimization, Data Augmentation, Computational Biology}

\vskip 0.3in
]



\printAffiliationsAndNotice{} 

\begin{abstract}
Self-supervised representation learning in computer vision relies heavily on hand-crafted image transformations to learn meaningful and invariant features. However few extensive explorations of the impact of transformation design have been conducted in the literature. In particular, the dependence of downstream performances to transformation design has been established, but not studied in depth. In this work, we explore this relationship, its impact on a domain other than natural images, and show that designing the transformations can be viewed as a form of supervision. First, we demonstrate that not only do transformations have an effect on downstream performance and relevance of clustering, but also that each category in a supervised dataset can be impacted in a different way. Following this, we explore the impact of transformation design on microscopy images, a domain where the difference between classes is more subtle and fuzzy than in natural images. In this case, we observe a greater impact on downstream tasks performances. Finally, we demonstrate that transformation design can be leveraged as a form of supervision, as careful selection of these by a domain expert can lead to a drastic increase in performance on a given downstream task.
\end{abstract}

\section{Introduction}
\label{introduction}

In Self-Supervised Representation Learning (SSRL), a model is trained to learn a common representation of two different transformations of the same image. The objective of SSRL is to benefit from training on a large unannotated dataset to obtain a representation that can be useful at solving downstream tasks for which one has a limited amount of annotated data. SSRL has become one of the main pillars of \textit{Deep Learning} based computer vision approaches~\cite{vicreg,dino,deepclusterv2,simclr,moco,byol,barlow_twins}, with performances coming close to, and sometimes going beyond supervised learning for some downstream tasks.

\begin{figure}[t]
\vskip 0.2in
\begin{center}
\centerline{\includegraphics[width=\columnwidth]{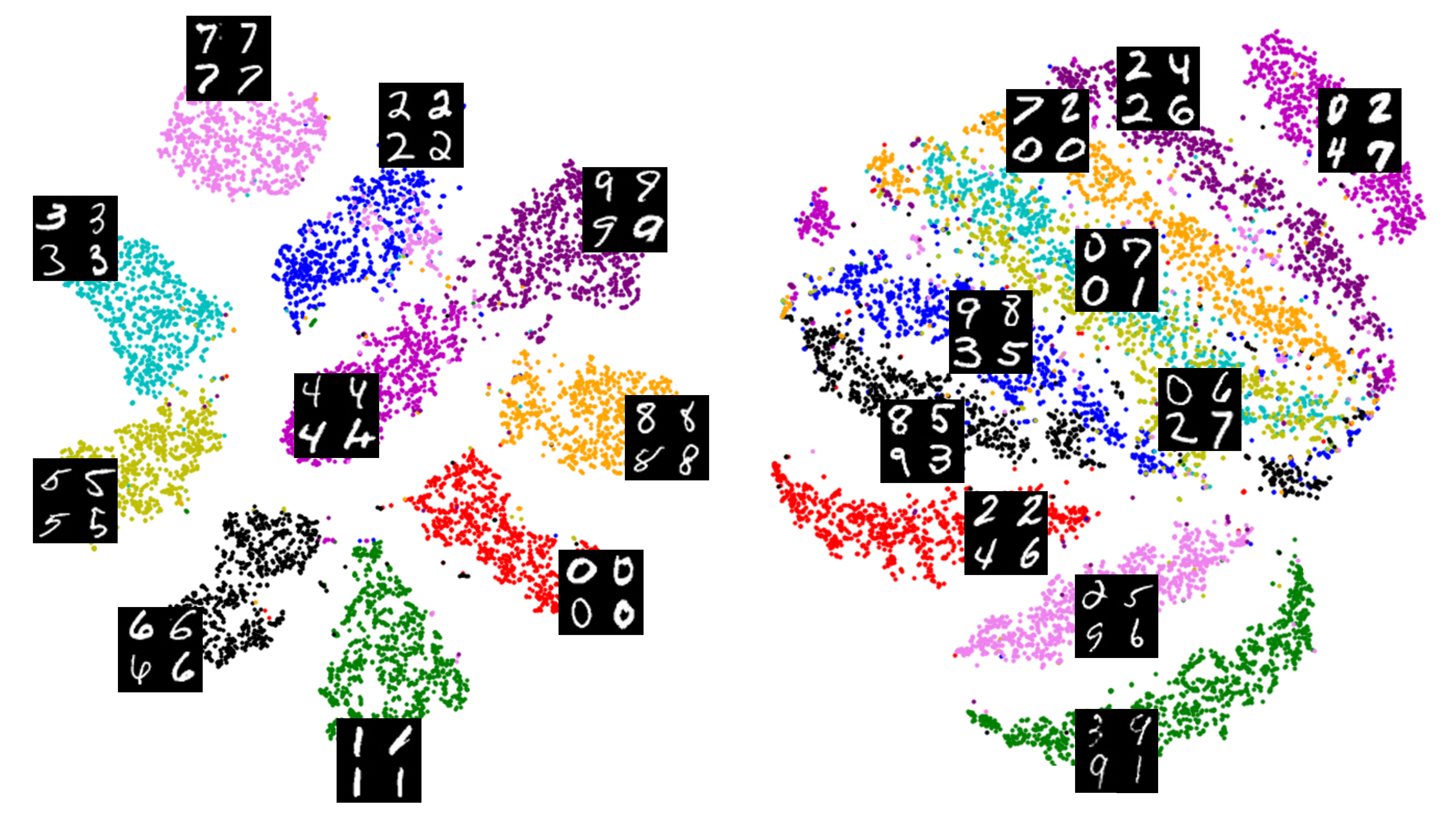}}
\caption{A t-SNE projection of the ten-class clustering of the MNIST dataset~\cite{mnist} performed on two representations obtained from two self-supervised trainings of the same model using MoCo V2~\cite{moco}, with the sole distinction being the selection of transformations employed. One representation retains information pertaining to the digit classes \textit{(left)}, achieved through padding, color inversion, rotation, and random cropping, while the other representation preserves information regarding the handwriting classes \textit{(right)}, achieved through vertical flips, rotation, random cropping. The selection of transformations dictates the features learnt during the training process, thus enabling the adaptation of a model for different downstream tasks.}
\label{fig:mnist_two_clusterings}
\end{center}
\vskip -0.2in
\end{figure}

SSRL relies heavily on combinations of random image transformations. These transformations are used to create distorted versions of the original image with the aim of keeping the semantic content invariant. With SSRL approaches producing overall good accuracy on downstream classification of natural images, hyper-parameter optimization of transformation parameters has added significant improvements to the overall performance of models~\cite{simclr}. However, further consequences of the choice of these augmentations have been only sporadically explored by the research community~\cite{byol,group_augment}, especially for other tasks~\cite{best_transformation_work} and in other domains~\cite{tutorial_contrastive}. It is therefore unclear to what extent this choice impacts the pretraining of models on a deeper level, as well as the performances on downstream tasks or in other domains.

Some important questions remain unanswered. Is the accuracy for an individual class contingent upon the choice of augmentation ? Can the variation of this choice increase one class accuracy at the expense of degrading another one? Are other downstream tasks, such as clustering, being affected by this choice ? What is the amplitude of these issues in domains other than natural images? 


In this paper, we report and analyze the outcomes of our experimentation to shed light on this subject. By examining the performance of various SSRL methods, while systematically altering the selection and magnitude of transformations, we analyze and quantify their ramifications on overall performance as well as on the class-level performance of models. We subsequently seek to observe the effects of substantially varying the structure of combinations of transformations on disparate downstream tasks, such as clustering and classification. We subsequently investigate the effects of varying selections transformations in SSRL methods when applied to microscopy images of cells, where distinctions between classes are far less discernible than with natural images, and then proceed to discuss potential avenues for improvement in the SSRL field.

Our contributions can be succinctly summarized as follows:
\begin{itemize}
\item We investigate the subtle effects of transformations on class-level performance. Our findings indicate that the transformations commonly employed in Self-Supervised Learning literature can impose penalties on certain classes while conferring advantages to others.
\item We demonstrate that the selection of specific combinations of transformations leads to the development of models that are optimized for specific downstream tasks while concurrently incurring penalties on other tasks.
\item We examine the implications of the choice of transformations for Self-Supervised Learning in the biological domain, where distinctions between classes are fuzzy and subtle. Our results reveal that the choice of transformations holds greater significance in this domain and suggest that the selection of transformations by domain experts leads to substantial enhancement of the quality of results.
\end{itemize}

\section{Related work}
\label{sec:related work}

\textbf{Self supervised representation learning (SSRL).} Contrastive learning approaches~\cite{simclr,vit_empirical_study,moco,decoupled_contrastive} have shown great success in avoiding trivial solutions in which all representations collapse into a constant point, by pushing the original image representation further away from representations of negative examples. These approaches follow the assumption that the augmentation distribution for each image has minimal inter-class overlap and significant intra-class overlap~\cite{large_scale_pretraining,theorical_overlap}. This dependence on contrastive examples has since been bypassed by non contrastive methods. The latter either have specially designed architectures~\cite{dino,simsiam,byol} or use regularization methods to constrain the representation in order to avoid the usage of negative examples~\cite{vicreg,whitening,vibcreg,neural_manifold,barlow_twins,vicregl}.
\\
\\
\noindent \textbf{Impact of image transformations on SSRL.} Compared to the supervised learning field, the choice and amplitude of transformations has not received much attention in the SSRL field~\cite{supervised_class_augments,autoaugment,dada,direct_differentiable_augments}. Studies such as~\cite{decoupling_features} and~\cite{hyper_sphere} analyzed in a more formal setting the manner in which augmentations decouple spurious features from dense noise in SSRL, and measures of alignment between features of transformed images. Some works~\cite{simclr,Geirhos_2020,byol,Perakis_2021} explored the effects of removing transformations on overall accuracy. Other works explored the effects of transformations by capturing information across each possible individual augmentation, and then merging the resulting latent spaces~\cite{tutorial_contrastive}, while some others suggested predicting intensities of individual augmentations in a semi-supervised context~\cite{optimized_transform}. However the latter approach is limited in practice as individual transformations taken alone were shown to be far less efficient than compositions~\cite{simclr}. An attempt was made to explore the underlying effect of the choice of transformation in~\cite{study_on_transforms}, one of the first works to discuss how certain transformations are preferable for defining a pretext task in self supervised learning. This study suggests that the best choice of transformations is a composition that distorts images enough so that they are different from all other images in the dataset. However favoring transformations that learn features specific to each image in the dataset should also degrade information shared by several images in a class, thus damaging model performance. Altogether, it seems that a good transformation distribution should maximize the intra-class overlap, while minimizing inter-class overlap~\cite{large_scale_pretraining,theorical_overlap}. Other works proposed a formalization to generalize the composition of transformations~\cite{formal_composition_analysis}, which, while not flexible, provided initial guidance to improve results in some contexts. This was followed by more recent works on the theoretical aspects of transformations,~\cite{content_from_style} that studied how SSRL with data augmentations identifies the invariant content partition of the representation,~\cite{transformation_generalization} that seeks to understand how image transformations improve the generalization aspect of SSRL methods, and~\cite{best_transformation_work} that proposes new hierarchical methods aiming to mitigate a few of the biases induced by the choice of transformations.
\\
\\
\noindent \textbf{Learning transformations for SSRL.}
A few studies showed that optimizing the transformation parameters can lead to a slight improvement in the overall performance in a low data annotation regime~\cite{optimized_transform,self_augment}. However, the demonstration is made for a specific downstream task that was known at SSRL training time, and optimal transformation parameters selected this way were shown not to be robust to slight change in architecture or task~\cite{inductive_biases}. Other works proposed optimizing the random sampling of augmentations by representing them as discrete groups, disregarding their amplitude~\cite{group_augment}, or through the retrieval of strongly augmented queries from a pool of instances~\cite{strong_augmentations}. Further research aimed to train a generative network to learn the distribution of transformation in the dataset through image-to-image translation, in order to then avoid these transformations at self supervised training time~\cite{gan_automate_transforms}. However, this type of optimization may easily collapse into trivial transformations. 
\\
\\
\noindent \textbf{Performance of SSRL on various domains and tasks.}
Evaluation of most SSRL works relies almost exclusively on accuracy of classification of natural images found in widely used datasets such as Cifar~\cite{cifar}, Imagenet~\cite{imagenet} or STL~\cite{STL_dataset}. This choice is largely motivated by the relative ease of interpretation and understanding of the results, as natural images can often be easily classified by eye. This, however, made these approaches hold potential biases concerning the type of data and tasks for which they could be efficiently used. It probably also has an impact on the choice and complexity of the selected transformations aiming at invariance: some transformations could manually be selected in natural images but this selection can be very challenging in domains where differences between classes are invisible. The latter was intuitively mentioned in some of the previously cited studies. Furthermore, the effect of the choice of transformation may have a stronger effect on domains and task where the representation is more thoroughly challenged. This is probably the case in botany and ornithology~\cite{tutorial_contrastive} but also in the medical domain~\cite{optimized_transform} or research in biology~\cite{limited_data_gan,Lamiable2022.06.16.496413,quality_control}.

\section{The choice of transformations is a subtle layer of weak supervision}
\label{sec:results}

In Section \ref{sec:interclass_bias}, we empirically investigate the ramifications of varying transformation intensities on the class-level accuracies of models trained using self-supervised learning techniques. Subsequently, in Section \ref{sec:mnist_clustering}, we conduct an examination in which we demonstrate how alternative selections of transformations can lead to the optimization of the model for distinct downstream tasks. In Section \ref{sec:diff_domain}, we delve into the manner in which this choice can impact downstream tasks on biological data, a domain where distinction between images is highly nuanced. This is followed by an empirical analysis in Section \ref{sec:incompatible transformations} that illustrates how the combination of transformations chosen by domain experts in the field of biology can significantly enhance the performance of models in SSRL.

\begin{figure*}[tb]
\vskip 0.2in
\begin{center}
\centerline{\includegraphics[width=\textwidth]{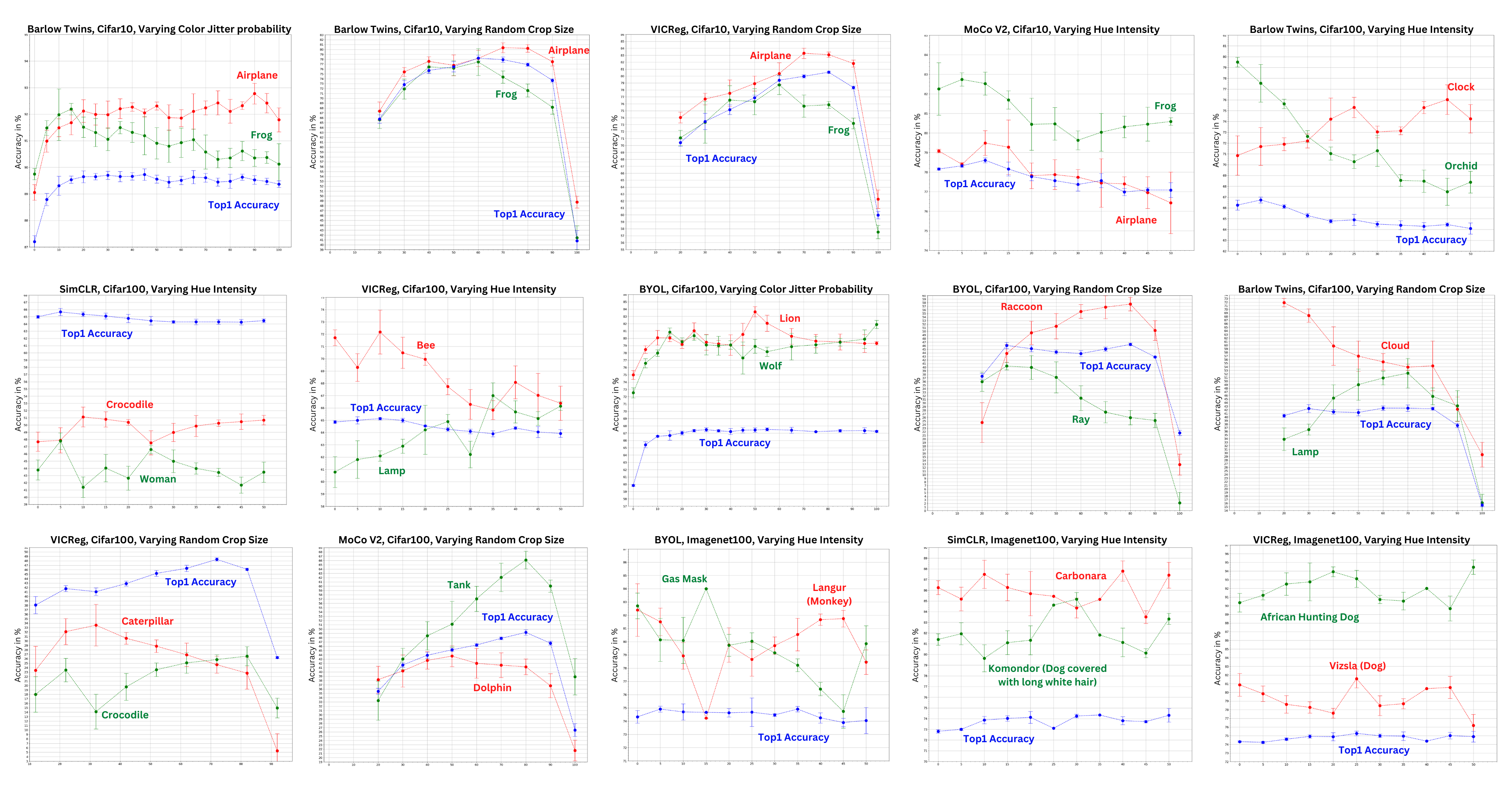}}
\caption{Inter-class accuracy results for Resnet18 architectures trained with various SSRL methods on the benchmark datasets Cifar10, Cifar100 and Imagenet100, as the parameters of different image transformations are varied. Each dot and associated error bar reflects the mean and standard deviation of three runs for Imagenet100 and five runs for Cifar, each with a different random seed. The results demonstrate that while overall accuracy remains relatively consistent across a range of transformation parameters, these transformations can have a subtle but significant impact on individual class performance, either favoring or penalizing specific classes.}
\label{fig:curves_classes}
\end{center}
\vskip -0.2in
\end{figure*}
 
\subsection{The choice of transformations induces inter-class bias}
\label{sec:interclass_bias}

In order to understand the ramifications of transformations on the performance of a model, we delve into the examination of the behavior of models that are trained with widely adopted SSRL techniques on the benchmark datasets of Cifar10, Cifar100~\cite{cifar} and Imagenet100~\cite{imagenet}, while systematically altering the magnitude and likelihood of the transformations. With a Resnet18 architecture as the backbone, we employ a fixed set of transformations, comprised of randomized cropping, chromatic perturbations, and randomized horizontal inversions. Subsequently, we uniformly sample a set of values for amplitude and probability for each transformation, in order to create a diverse range of test conditions. Each training is repeated a number of times (three for Imagenet and five for Cifar), with distinct seed values, and the mean and standard deviation of accuracy, measured through linear evaluation, is computed over these five trainings for each method and each transformation value. All model training parameters, as well as the training process, are available in Supplementary Materials \ref{sec_sup:interclass_bias}.


\begin{figure}[h]
\vskip 0.2in
\begin{center}
\centerline{\includegraphics[width=\columnwidth]{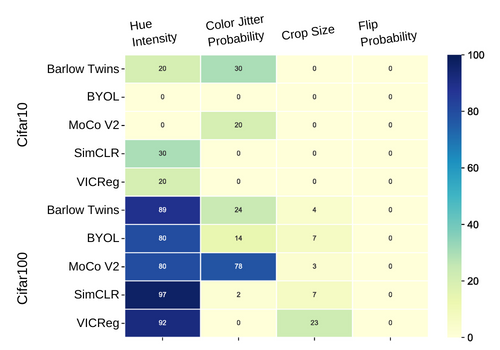}}
\caption{A comparison of the ratio of classes that exhibit negative correlations with one another, in relation to the total number of classes within Cifar10 and Cifar100. We limit our analysis to Pearson, Spearman and Kendall correlation coefficients lower than -0.3 and a p-value lower than 0.05. We observe a greater proportion of negatively correlated classes within Cifar100, due to the increased number of classes that may overlap when varying color jitter. It is important to note that different transformations can affect performance differently as they strive to preserve the features that define a class across the original image and its transformed versions. However, this may also inadvertently compromise information specific to a particular class, while favoring the information of another class.}
\label{fig:correlation_matrix}
\end{center}
\vskip -0.2in
\end{figure}

As depicted in Figure \ref{fig:curves_classes}, we observe minimal fluctuation in the overall accuracy of each model as we slightly alter any of the transformations examined. This stands in stark contrast to the class-level accuracies observed, in which we discern significant variation in the accuracy value for many classes, as we vary the parameters of transformations, hinting at a greater impact of variations in transformation parameters on the class-level. Through the same figure, it becomes apparent that a number of classes exhibit distinct, and at times, entirely antithetical behaviors to each other within certain ranges of a transformation parameter. In the context of the datasets under scrutiny, this engenders a bias in the conventional training process of models, which either randomly samples transformation parameters or relies on hyperparameter optimization on overall accuracy to determine optimal parameters. This bias manifests itself in the manner in which choosing specific transformation parameters would impose a penalty on certain classes while favoring others. This is demonstrated in Figure \ref{fig:curves_classes} by the variation in accuracy of the Caterpillar and Crocodile classes for a model trained using VICReg~\cite{vicreg}, as the crop size is varied. The reported accuracies reveal that a small crop size benefits the Caterpillar class, as it encourages the model to learn features specific to its texture, while the Crocodile class is negatively impacted in the same range. These results suggest that a given choice of transformation probability or intensity has a direct effect on class level accuracies that is not necessarily visible through overall accuracy.


In order to gain deeper understanding of the inter-class bias observed in our previous analysis, we aim to further investigate the extent to which this phenomenon impacts the performance of models trained with self-supervised learning techniques. By quantitatively assessing the correlation scores between class-level accuracies obtained under different transformation parameters, we aim to clearly and unequivocally measure the prevalence of this bias in self-supervised learning methods. More specifically, a negative correlation score between the accuracy of two classes in response to varying a given transformation would indicate opposing behaviors for those classes. Despite its limitations, such as the inability to quantify the extent of bias and the potential for bias to manifest in specific ranges while remaining positively correlated in others (See Lion/Wolf pair in Figure \ref{fig:curves_classes}), making it difficult to detect, this measure can still provide a preliminary understanding of the degree of inter-class bias. To this end, we conduct a series of experiments utilizing a ResNet encoder on the benchmark datasets of Cifar10 and Cifar100~\cite{cifar}. We employ a diverse set of state-of-the-art self-supervised approaches: Barlow Twins~\cite{barlow_twins}, MoCov2~\cite{moco}, BYOL~\cite{byol}, SimCLR~\cite{simclr}, and VICReg~\cite{vicreg}, and use the same fixed set of transformations as in our previous analysis depicted in Figure \ref{fig:curves_classes}. We systematically vary the intensity of the hue, the probability of color jitter, the size of the random crop, and the probability of horizontal inversion through 20 uniformly sampled values for each, and repeat each training five times with distinct seed values. We compute the Pearson, Kendall and Spearman correlation coefficients for each pair of classes with respect to a given transformation parameter, as well as their respective p-values, and define class pairs with opposite behaviors as those with at least one negative correlation score of the three measured correlations lower than -0.3 and a p-value lower than 0.05. We then measure the ratio of classes with at least one opposite behavior to another class, compared to the total number of classes, in order to understand the extent of inter-class bias for a given transformation, method, and dataset.


Our findings, as represented in Figure \ref{fig:correlation_matrix}, indicate that the extent of inter-class bias for the self-supervised learning methods of interest varies among different transformations. This variability is primarily due to the fact that while these transformations aim to preserve the features that define a class across the original image and its transformed versions, they can also inadvertently compromise information specific to a particular class, while favoring the information of another class. It is noteworthy that for Cifar100, which comprises a vast array of natural image classes, we observe a considerable number of classes that exhibit inter-class bias when manipulating hue intensity. This is most likely a consequence of the high degree of overlap between classes when altering the colors within each class. Therefore, the selection of specific transformations and their associated parameters must be carefully tailored to the information that we aim to preserve within our classes, which constitutes a form of weak supervision in the choice of which classes to prioritize over others.

\subsection{The choice of transformations directly affects downstream tasks}
\label{sec:mnist_clustering}

\begin{figure*}[tb]
\vskip 0.2in
\begin{center}
\centerline{\includegraphics[width=\textwidth]{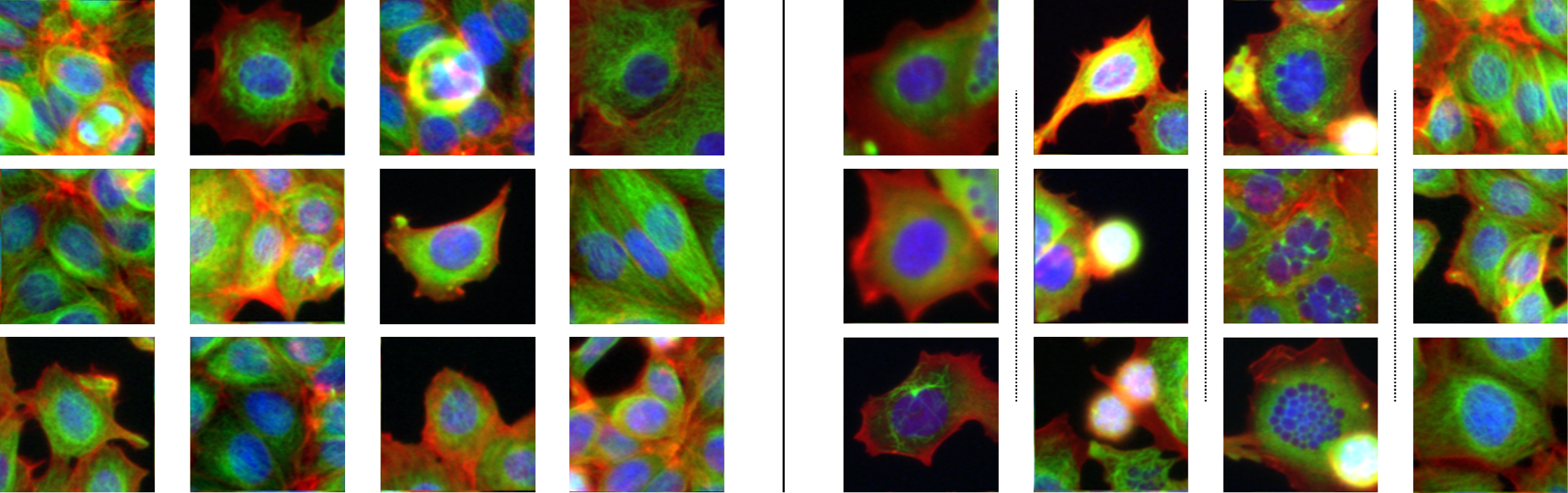}}
\caption{As a result of their unique genetic expression and surrounding environment, single cells \textit{(left - untreated cells)} are already substantially dissimilar within a given condition, complicating the identification and measurement of various perturbations. The figure illustrates \textit{(right - cells treated with high concentration of Nocodazole)} that a given treatment can produce four morphological responses (each column), with one of which can appear similar to untreated cells (most right). The majority of treatments at lower concentrations yield phenotypes that are indistinguishable by the naked eye from untreated cells (data not shown). The images of this dataset are cropped such that there is always a cell at the center.}
\label{fig:phenotypes_example}
\end{center}
\vskip -0.2in
\end{figure*}


Examining the impact of transformations on downstream tasks, including variations in transformation parameters and compositions, is crucial for a complete understanding of their influence. To this end, we direct our investigation towards the representations that can be attained by training a VGG11 encoder architecture~\cite{vgg_model} with a MoCo v2 approach~\cite{moco} on the MNIST benchmark dataset~\cite{mnist}. Our study delves into the effects of different sets of compositions on the nature of the information embedded within the representation and its correlation to downstream tasks. For each representation obtained, we conduct a K-Means clustering~\cite{kmeans} with ten clusters, and a linear evaluation utilizing the digit labels. In order to measure the efficacy of the clustering, we employ the Silhouette score~\cite{silhouette} to assess the quality of the distribution of data within and between clusters without the use of labels. Additionally, we use the Adjusted Mutual Information score (AMI)~\cite{ami}, which is a modified Mutual Information score that mitigates the influence of chance in clustering evaluations in comparison to clustering accuracy~\cite{clustering_accuracy}, by determining the level of agreement between the groupings of each cluster and the groupings of the classes in the ground truth. A more in-depth examination of the AMI score can be found in Supplementary materials, Section \ref{sec_sup:ami_score_explanation}.

\begin{table}[h]
\caption{The measured metrics for the clustering and linear evaluation of two models trained with two distinct sets of transformations on the MNIST dataset~\cite{mnist}. The first set of transformations comprises padding, color inversion, rotation and random cropping, while the second set is composed of vertical flips, rotation, random cropping and random erasing. Top1 Accuracy is computed through Linear Evaluation. It is noteworthy that the quality of the clusters, as measured by the Silhouette score~\cite{silhouette}, is relatively good for the second set, despite the clustering resulting from it failing to accurately capture the digit clusters as shown by the AMI score.}
\label{tab:mnist_table}
\vskip 0.15in
\begin{center}
\begin{small}
\begin{sc}
\resizebox{\columnwidth}{!}{%
\begin{tabular}{lcccr}
\toprule
Transformation sets & Silhouette & AMI & Top1 Acc   \\
\midrule
First Set & 0.87$\pm$ 0.03& 0.83$\pm$ 0.02 & 99.6$\pm$ 0.21 \\
Second Set & 0.66$\pm$ 0.08& 0.37$\pm$ 0.06 & 62.1$\pm$ 2.94 \\
\bottomrule
\end{tabular}
}
\end{sc}
\end{small}
\end{center}
\vskip -0.1in
\end{table}


For our trainings, we employ two sets of transformations. The first set, comprised of padding, color inversion, rotation, and random cropping, aims to maximize the intra-class overlap of digits. The second set, which includes vertical flips, rotation, random cropping, and random erasing, is utilized to investigate the representation resulting from the destruction of digit information. Each training iteration is repeated five times with distinct seeds, and the average of their scores is computed. As depicted in Table \ref{tab:mnist_table}, there is a decrease in both the AMI score and the top1 accuracy score for the training with the second set of transformations compared to the first set. This is to be expected as the second set of transformations is highly destructive to the digit information, resulting in a decline in performance that cannot be fully remedied, even with the supervised training used in linear evaluation. However, we observe that there is a slight decrease in the silhouette score compared to the decrease in the AMI score, with the score indicating that the clusters achieved from the representations of the second set of transformations are well-separated. As demonstrated in Figure \ref{fig:mnist_two_clusterings}, the clusters resulting from the two representations are distinguishable. The clusters resulting from the second set of transformations encapsulate handwriting information, such as line thickness and writing flow, forming handwriting classes for which the transformations maximized the intra-class overlap and minimized the inter-class overlap. This experiment implies that we can selectively supervise the specific image features that we aim to distinguish through a deliberate or unconscious selection of the transformations used during training, which would enhance the performance of the trained model for the associated downstream task.

\subsection{The relevance of transformations is domain dependant}
\label{sec:diff_domain}

With the objective of exploring the amplitude of the effect of transformations on other domains, we perform experiments on microscopy images of cells under two conditions (untreated vs treated with a compound) available from BBBC021v1~\cite{BBBC021}, a dataset from the Broad Bioimage Benchmark Collection~\cite{Ljosa2012}. Cells in their natural state can vary widely in appearance even when in the same condition, offering a more challenging context for SSRL (Figure \ref{fig:phenotypes_example}). In order to observe the effect of transformations in such a domain, we preprocess these microscopy images by detecting all cell nuclei and extracting an 196x196 pixels image around each of them. We focus our study on three main compounds : Nocodazole, Cytochalasin B and Taxol. Technical details of the dataset used and the data preprocessing performed can be found in Supplementary Material \ref{sec_sup:biology_context}. We use a VGG13~\cite{vgg_model} encoder architecture and MoCov2~\cite{moco} as the self supervised approach, and run two separate trainings of the model from scratch for each compound, each of the two trainings with a different composition of transformations for invariance, repeated five times with distinct seeds. We then perform a K-Means~\cite{kmeans} clustering (k=2) on the inferred test set embeddings and compute the Adjusted Mutual Information score (AMI)~\cite{ami} with respect to the ground truth compound labels of the compounds data subsets (untreated vs treated with Nocodazole, untreated vs treated with Cytochalasin B, untreated vs treated with Taxol).

\begin{table}[h]
\caption{The results of the adjusted mutual information score~\cite{ami} obtained for two sets of transformations, through the mean of five training runs for each, compared to each other and to the AMI score achieved through the representations of a pre-trained Resnet 101 on the dataset subset containing Nocodazole. The selection of Resnet width is studied in Supplementary Materials \ref{sec_sup:biological_clustering}. Both sets of transformations comprise affine transformations and color jitter, with the first set incorporating an additional random cropping, and resulting in a mediocre AMI score, and the second set utilizing random rotations and resulting in a significantly higher score. The significance of selecting the appropriate transformations to learn optimal representations in self-supervised learning is significantly greater in a domain where the features separating images are highly subtle.}
\label{tab:benchmark_phenostyle}
\vskip 0.15in
\begin{center}
\begin{small}
\begin{sc}
\resizebox{\columnwidth}{!}{%
\begin{tabular}{lcccr}
\toprule
 & Nocodazole & Cytochalasin B & Taxol   \\
\midrule
First Set & 0.19$\pm$0.01 & 0.27$\pm$0.02 & 0.16$\pm$0.05  \\
Second Set & 0.37$\pm$0.03 & 0.45$\pm$0.01 & 0.38$\pm$0.01  \\
Resnet101 & 0.39& 0.57 & 0.43\\
\bottomrule
\end{tabular}
}
\end{sc}
\end{small}
\end{center}
\vskip -0.1in
\end{table}

For each composition of transformations explored, we repeat the training five times with different seeds, and compute the average and standard deviation of the AMI score (technical details of the  the model training can be found in Supplementary material Section  \ref{sec_sup:biological_clustering}). Table \ref{tab:benchmark_phenostyle} displays the AMI scores achieved by using two different compositions of transformations in training, compared to the AMI score of a clustering on representations achieved with a pretrained Resnet101. By replacing slight random cropping and resizing, a transformation used in most existing self supervised approaches, with very strong random rotations of the image, we report a significantly higher mean AMI score, which shows that the model using random rotations is able to learn representations that better separate untreated from compound treated cells, including cells displaying subtle differences unnoticeable by the naked eye. Inversely, random cropping was destructive to the representations in this case, as the cropped images can miss out on relevant features in the sides of the cell. Compared to the slight effects of transformations reported on overall accuracy on Cifar10 and Imagenet100 in section \ref{sec:interclass_bias}, the significantly higher effect observed on this type of data suggests that transformations can have a higher impact on learning effective representations from images holding subtle differences between classes, and come to achieve similar performances to models pretrained with supervision.

\begin{figure}[h!]
\vskip 0.2in
\begin{center}
\centerline{\includegraphics[width=\columnwidth]{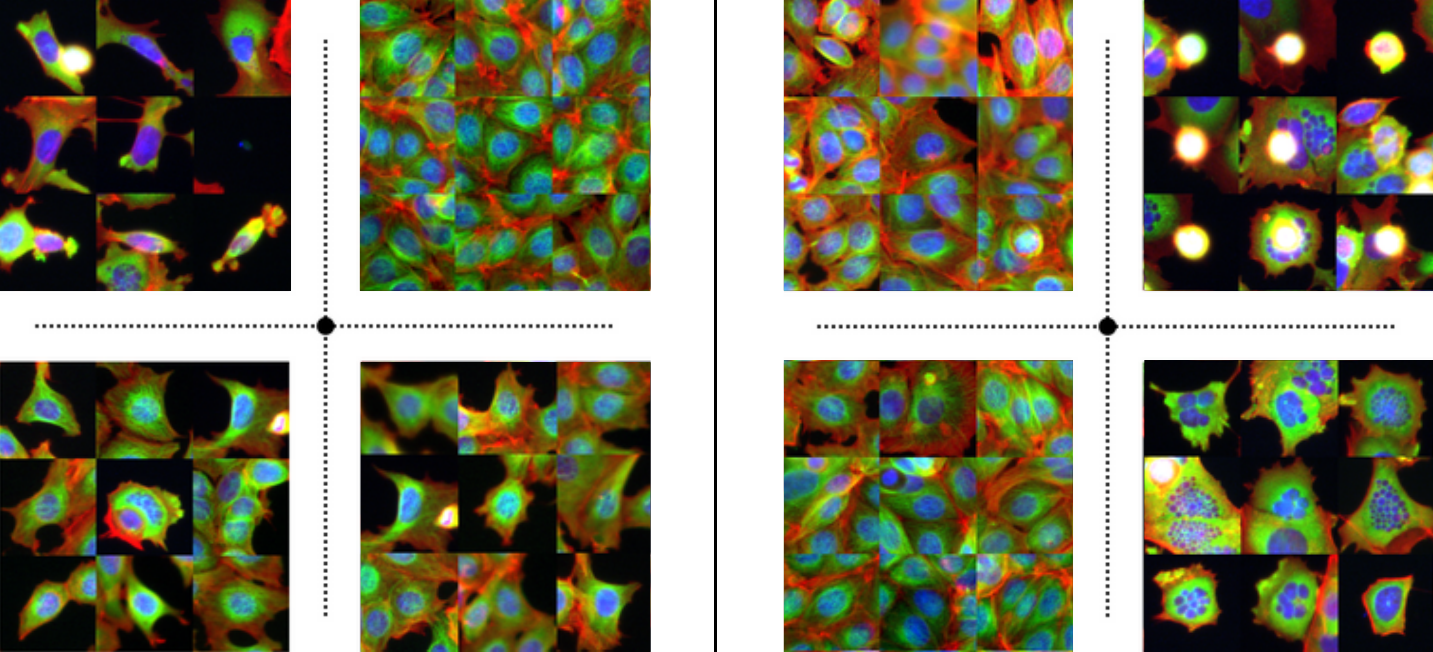}}
\caption{An illustration of the various K-Means (k=4) clustering results on the data subset of Nocodazole defined in Supplementary Material \ref{sec_sup:biology_context}, utilizing different combinations of augmentations, with the objective of separating the distinct morphological reactions of the cells into different clusters. The sampled images are the closest images to the centroid of the cluster using Euclidean distance on the representations. The clusters on the left predominantly take into consideration the number of cells present in each image, and result from training the model with a combination of color jitter, flips, rotation, affine transformation, and random cropping. The clusters on the right take into account some of the phenotypes under examination, and result from training the model with a combination of rotations, center cropping, color jitter, and flips. The parameters of each transformation are detailed in Supplementary Material \ref{sec_sup:biological_clustering}.}
\label{fig:clustering_phenotype}
\end{center}
\vskip -0.2in
\end{figure}

Beyond clustering into two conditions, we wonder what combination of transformations could lead to a proper clustering of cell phenotypes (or morphology). We explore different compositions of transformations in additional experiments with the same VGG13~\cite{vgg_model} architecture and MoCov2~\cite{moco} loss function. We then apply K-Means~\cite{kmeans} clustering (k=4) on the representations obtained from the test set. As observed in Section \ref{sec:mnist_clustering}, different compositions of transformations can lead to very different clustering results. This is further confirmed in the microscopy domain for Nocodazole (Figure \ref{fig:clustering_phenotype}), as we observe that the composition of color jitter, flips, rotation, affine, in addition to random crops, would result in clustering images by the number and size of cells, rather than by morphological features (\textit{Figure \ref{fig:clustering_phenotype} left}). We perform a training where affine transform and random crop were replaced by a center crop that preserves 50\% of the image around the central cell. The latter resulted in four clusters where two out of the three cell phenotypes were detected. However, it also had the effect of splitting untreated cells into two different clusters (\textit{Figure \ref{fig:clustering_phenotype} right}). Altogether, engineering a combination of transformations represents a somewhat weak supervision that can become a silent bias or, alternatively, can be leveraged as a powerful tool to achieve a desirable result.

\subsection{Improving representations requires a choice of transformations based on domain expertise}
\label{sec:incompatible transformations}

\begin{figure*}[tb]
\vskip 0.2in
\begin{center}
\centerline{\includegraphics[width=\textwidth]{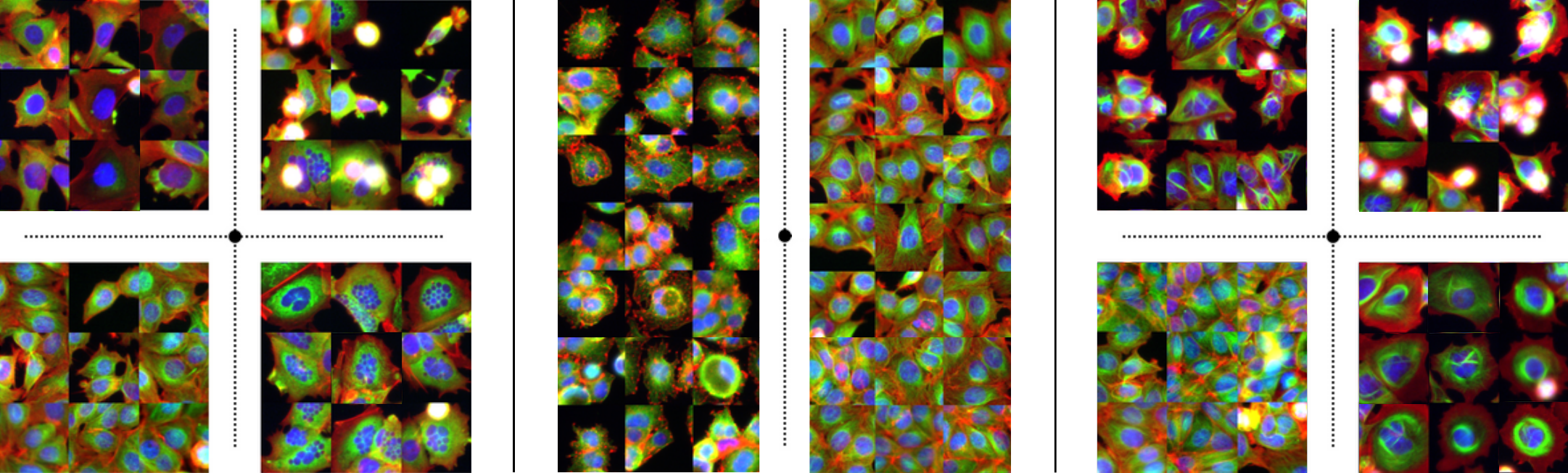}}
\caption{The clustering results achieved through the utilization of two MoCo v2 losses~\cite{moco}, each with a distinct set of transformations, on the Nocodazole \textit{(left)}, Cytochalasin B \textit{(middle)}, and Taxol \textit{(right)} data subsets. One loss employs color jitter, flips, rotation, affine transformation, and random cropping, while the other utilizes rotations, center cropping, color jitter, and flips. The clustering results demonstrate that the phenotypes of each subset are clearly separated and represented in each cluster, as evidenced by the images closest to its centroïd.}
\label{fig:correct_phenotypes}
\end{center}
\vskip -0.2in
\end{figure*}

In the previous section, in order to achieve a clustering that extracts the three main cell morphologies produced by the Nocodazole treatment, we hypothesize that a focus on the center of the image (which is also necessarily the center of a cell given the preprocessing step), coupled with a focus on the relationship of the center cell to its surroundings were both necessary. We test this hypothesis by performing distinct trainings that combine both sets of transformations used in Figure \ref{fig:clustering_phenotype} on the three compound data subsets : Nocodazole, Taxol and Cytochalasin B. Doing so produces some incompatible combinations of transformations, for instance, it brings together a random crop followed by a center crop, we keep both compositions separated in two independent Mocov2 losses. We then train a model to minimize the weighted sum of these two losses on each of the compound data subsets, and perform a K-Means clustering (k=4) on the resulting representations for the Nocodazole and Taxol subsets, and a K-Means clustering (k=2) on the resulting representation for the Cytochalasin B subset.

\begin{table}[h]
\caption{A comparison of the AMI score of the clusterings achieved through a K-Means (k=2) over the representations of models trained through our combination of two MoCo v2 losses and the representations of a pre-trained Resnet101 encoder, on the biological data subsets of Nocodazole, Cytochalasin B, and Taxol. One loss employs color jitter, flips, rotation, affine transformation, and random cropping, while the other utilizes rotations, center cropping, color jitter, and flips. Through conscious selection of a combination of different sets of transformations that are adapted to the task by a domain expert, we surpass the performance of pre-trained Resnets on clustering through SSRL pre-training.}
\label{tab:compound_table}
\vskip 0.15in
\begin{center}
\begin{small}
\begin{sc}
\resizebox{\columnwidth}{!}{%
\begin{tabular}{lcccr}
\toprule
 & Nocodazole & Cytochalasin B & Taxol   \\
\midrule
Ours & 0.51 & 0.66& 0.52  \\
Resnet101 & 0.39& 0.57 & 0.43\\
\bottomrule
\end{tabular}
}
\end{sc}
\end{small}
\end{center}
\vskip -0.1in
\end{table}

In the subsequent results in Figure \ref{fig:correct_phenotypes} we can observe that we successfully separate all cell phenotypes obtained after each of the three compound treatments, from each other and from untreated cells, which validates our hypothesis. By  performing another K-Means (k=2) on the same representations, we also report AMI scores for the task of separating the compound-treated cells from untreated cells, that can be considered quite high in this context where treated cells can look like untreated cells, vastly surpassing previous trainings with separate compositions of transformations in Table \ref{tab:compound_table}, as well as surpassing the performance of pretrained models. This suggests that more complex transformation manipulations, beyond modifying a single transformation parameter, can yield a better representation that significantly improves results on some specific tasks. However, identifying the right combination definitely requires domain expertise.

\section{Conclusion}
\label{sec:concl}

In this work, we performed experiments showing the impact of the choice, amplitude and combination of the transformations on the effectiveness of the self supervised representation that is learnt. We showed that the choice of transformation induces an inter-class bias but can also strongly affect other downstream tasks such as clustering, leading to unexpected results. We also demonstrated that the consequence of the choice of transformations can be very amplified in other domains where the difference between class is fuzzy or less visually identifiable. Altogether, some of the results can be understood somewhat intuitively. If one erases color from a car dataset, a deep network might not find enough correlated information to be able to classify cars on the basis of their original color. Thus the question: what is a good representation? The correct answer is that it depends on the downstream task. While it is what originally SSRL aimed to avoid, we showed, nonetheless, that choosing a rightful combination of transformations assembled in a non straightforward way by a domain expert can lead to very effective results.

\section{Acknowledgments}
\label{sec:acknowledgments}
This work was supported by ANR–10–LABX–54 MEMOLIFE and ANR–10 IDEX 0001–02 PSL* Université Paris and was granted access to the HPC resources of IDRIS under the allocation 2020-AD011011495 made by GENCI.

\bibliography{example_paper}
\bibliographystyle{icml2023}

\newpage
\appendix
\onecolumn
\section{Datasets}
\label{sec_sup:biology_context}

We perform the experiments mentioned in Section \ref{sec:diff_domain} on microscopy images available from BBBC021v1~\cite{BBBC021}, a dataset from the Broad Bioimage Benchmark Collection~\cite{Ljosa2012}. This dataset is composed of breast cancer cells treated for 24 hours with 113 small molecules at eight concentrations, with the top concentration being different for many of the compounds through a selection from the literature. Throughout the quality control process, images containing artifacts or with out of focus cells were deleted, and the final dataset totalled into 13,200 fields of view, imaged in three channels, each field composed of thousands of cells. In the cells making up the dataset, twelve different primary morphological reactions from the compound-concentrations were identified, with only six identified visually, while the remainder were defined based on the literature, as the differences between some morphological reactions were very subtle. We perform a simple cell detection in each field of view, in order to crop cells centered in (196x196px) images. We then filter the images by compound-concentration, and keep images treated by Nocodazole at its 4 highest concentrations. These compound-concentrations result in 4 morphological cell reactions, for which we don't have individual labels for each cell image. We sample the same number of images from views of untreated cells, totalling into a final data subset of 3500 images, which we split into 70\% training data, 10\% validation data and 20\% test data. We repeat the same process for the compounds Taxol and Cytochalasin B, each containing 4 and 2 morphological reactions respectively, to result in data subsets of sizes 1900 and 2300 images, respectively.

\section{Model training}
\label{sec_sup:model_training}

\subsection{Study of inter-class bias in self supervised classification}
\label{sec_sup:interclass_bias}

For results in Section \ref{sec:interclass_bias} and Table \ref{tab:model_stats_cifar_0_1}, we run several trainings of 12 SSRL methods~~\cite{vicreg,deepclusterv2,swav,simclr,simsiam,moco,nnclr,byol,vibcreg,barlow_twins,ressl} on Cifar10 and Cifar100~\cite{cifar} with a Resnet18 architecture, with a pretraining of 1000 epochs without labels. We use the same transformations with similar parameters on all approaches, namely a 0.4 maximal brightness intensity, 0.4 maximum contrast intensity, 0.2 maximum saturation intensity, all with a fixed probability of 80\%. With a maximal hue intensity of 0.5, we vary the hue probability of application between 0\% and 100\% by uniformly sampling 20 probability values in this range. We use stochastic gradient descent as optimization strategy for all approaches, and through a line search hyperparameter optimization, we use a batch size of 512 for all approaches except Dino~\cite{dino} and Vicreg~\cite{vicreg} for which we use a batch size of 256. Following the hyperparameters used in the literature of each approach, we use a projector with a 256 output dimension for most methods, except for Barlow Twins~\cite{barlow_twins}, Simsiam~\cite{simsiam}, Vicreg~\cite{vicreg} and Vibcreg~\cite{vibcreg}, for which we use a projector with a 2048 output dimension, and DeepclusterV2~\cite{deepclusterv2} and Swav~\cite{swav} for which we use a projector with a 128 output dimension. For some momentum based methods (Byol~\cite{byol}, MocoV2+~\cite{moco}, NNbyol~\cite{nnclr} and Ressl~\cite{ressl}), we use a base Tau momentum of 0.99 and use a base Tau momentum of 0.9995 for Dino~\cite{dino}. For Mocov2+~\cite{moco}, NNCLR~\cite{nnclr} and SimCLR~\cite{simclr}, we use a temperature of 0.2.

We train the Barlow Twins~\cite{barlow_twins} based model with a learning rate of 0.3 and a weight decay of \(10^{-4}\), and Byol~\cite{byol} as well as NNByol~\cite{nnclr} with a learning rate of 1.0 and a weight decay of \(10^{-5}\). We train DeepclusterV2~\cite{deepclusterv2} with a learning rate of 0.6, 11 warmup epochs, a weight decay of \(10^{-6}\), and 3000 prototypes. We train Dino with a learning rate of 0.3, a weight decay of \(10^{-4}\), and 4096 prototypes, while we train MocoV2+~\cite{moco} with a learning rate of 0.3, a weight decay of \(10^{-4}\) and a queue size of 32768. For NNclr~\cite{nnclr}, we use a learning rate of 0.4, a weight decay of \(10^{-5}\), and a queue size of 65536. We train ReSSL~\cite{ressl} with a learning rate of 0.05, and a weight decay of \(10^{-4}\), while we train SimCLR~\cite{simclr} with a learning rate of 0.4, and a weight decay of \(10^{-5}\). For Simsiam~\cite{swav}, we use a learning rate of 0.5, and a weight decay of \(10^{-5}\), and use for Swav~\cite{swav} a learning rate of 0.6, a weight decay of \(10^{-6}\), a queue size of 3840, and 3000 prototypes. We use for Vicreg~\cite{vicreg} and Vibcreg~\cite{vibcreg} a learning rate of 0.3, a weight decay of \(10^{-4}\), an invariance loss coefficient of 25, and a variance loss coefficient of 25. We use a covariance loss coefficient of 1.0 for Vicreg~\cite{vicreg} and a covariance loss coefficient of 200 for Vibcreg~\cite{vibcreg}. We perform linear evaluation after each pretraining for all methods, through freezing the weights of the encoder and training a classifier for 100 epochs. We use 5 different global seeds (5, 6, 7, 8, 9) for each hue intensity value, and compute the mean top1 accuracy resulting from the linear evaluation, using each of the 5 different experiences. Each training run was made on a single V100 GPU. On ImageNet100~\cite{imagenet}, we train a Resnet18 encoder with BYOL~\cite{byol}, MoCo V2~\cite{moco}, VICReg~\cite{vicreg} and SimCLR~\cite{simclr}, using a batch size of 128 for 400 epochs. We use a learning rate of 0.3 and a weight decay of \(10^{-4}\) for MoCo V2 and VICReg, and a weight decay of \(10^{-5}\)  and learning rates of 0.4 and 0.45 for SimCLR and BYOL respectively. We repeat each experience three times with three global seeds (5,6 and 7)  and compute its mean and standard deviation.

For the results in Figures \ref{fig:curves_classes} and \ref{fig:curves_classes}, we reuse the same training hyperparameters for Barlow Twins~\cite{barlow_twins}, Moco V2~\cite{moco}, BYOL~\cite{byol}, SimCLR~\cite{simclr} and Vicreg~\cite{vicreg}, and uniformly sample 10 values in the range of [0;0.5] for the maximal hue intensity, with a fixed 80\% probability. We run different experiments for the 5 global seeds for each hue intensity, and compute their mean and standard deviation. We perform the same process while fixing maximal hue intensity to 0.1, and varying its probability by uniformly sampling 20 probability values in the range [0;100]. We repeat a similar process for the random cropping and horizontal flips, by sampling 8 values uniformly in the range of [20;100] of the size ratio to keep of the image, and sampling 20 values uniformly in the range of [0;100] for the probability of application of horizontal flips.

\subsection{MNIST Clustering}
\label{sec_sup:mnist_clustering}

For the displayed clustering results in Figure \ref{fig:mnist_two_clusterings}, we use a VGG11~\cite{vgg_model} architecture, with a projector of 128 output dimension, trained using a MocoV2+~\cite{moco} loss function on Mnist~\cite{mnist} for 250 epochs. We use an Adam optimizer, a queue size of 1024, and a batch size of 32. We set temperature at 0.07, learning rate at 0.001, and weight decay at 0.0001. We run two trainings with two separate sets of compositions of transformations, each run on a single V100 GPU, and perform a Kmeans (K=10) clustering on the resulting representations of the test set. For the digit clustering, we use a composition of transformations composed of a padding of 10\% to 40\% of the image size, color inversion, rotation with a maximal angle of 25°, and random crop with a scale in the range of [0.5;0.9] of the image, and then a resizing of the image to 32x32 pixels. For the handwriting flow clustering, we use a composition of transformations composed of horizontal \& vertical flips with an application probability of 50\% each, rotations with a maximal angle of 180°, random crop with a scale in the range of [0.9;1.1], and random erasing of patches of the image, with a scale in the range of [0.02;0.3] of the image and a probability of 50\%. We perform linear evaluation by training classifiers on the frozen representations of the trained models, in order to predict the digit class, and evaluate using the top1 accuracy score.

\subsection{Clustering evaluation with the AMI Score}
\label{sec_sup:ami_score_explanation}

We use the adjusted mutual information (AMI)~\cite{ami} in Sections \ref{sec:diff_domain} and \ref{sec:incompatible transformations} to evaluate clustering quality, and to measure the similarity between two clusterings. It is a value that ranges from 0 to 1, where a higher value indicates a higher degree of similarity between the two clusterings. This score holds an advantage over clustering accuracy~\cite{clustering_accuracy} in one main aspect, being that the clustering accuracy only measures how well the clusters match the labels of the true clusters, and does not take into account the structure within the clusters, such as heterogeneity of some of the clusters. This is unlike the AMI score, which takes into account both the structure between the clusters and the structure within the clusters, by measuring the "agreement" between the groupings of a predicted cluster and the groupings of the true cluster. If both clusterings agree on most of the groupings, then the AMI score will be high, and inversely low if they do not. 

The AMI score can be computed with the formula :

$$ AMI(X,Y) = \frac{MI(X,Y) - E(MI(X,Y))}{max(H(X), H(Y)) - E(MI(X,Y))} $$

Where $MI(X,Y)$ is the mutual information between the two clusterings, $E(MI(X,Y))$ is the expected mutual information between the two clusterings, $H(X)$ is the entropy of the clustering $X$, and $H(Y)$ is the entropy of the clustering $Y$. Mutual information (MI) is a measure of the amount of information that one variable contains about another variable. In the context of AMI, the two variables are the clusterings $X$ and $Y$. $MI(X,Y)$ is a measure of to what extent the two clusterings are related to each other. Entropy is a measure of the amount of uncertainty in a random variable. In the context of AMI, the entropy of a clustering ($H(X)$ or $H(Y)$) is a measure of how much uncertainty exists within the clustering. Expected mutual information ($E(MI(X,Y))$) is the average mutual information between the two clusterings, assuming that the two clusterings are independent.

The adjusted mutual information (AMI) is calculated by first subtracting the expected mutual information ($E(MI(X,Y))$) from the actual mutual information ($MI(X,Y)$). This results in a measure of of the extent of the relationship between the two clusterings beyond what would be expected by chance. This value is then divided by the difference between the maximum possible entropy ($max(H(X), H(Y))$) and the expected mutual information ($E(MI(X,Y))$). Normalization of the result is achieved through this process, ensuring that it is always between 0 and 1. Figure \ref{fig:resnet} shows the results of a clustering achieved on Nocodazole vs untreated cells, with the AMI score computed after randomisation of the ground truth labels, in contrast to clustering results achieved without randomizing the labels.

\subsection{Cellular Clustering}
\label{sec_sup:biological_clustering}

For the results in Sections \ref{sec:diff_domain}, \ref{sec:incompatible transformations}, we use a VGG13~\cite{vgg_model} architecture, trained using a MocoV2+~\cite{moco} loss function on the data subsets of the microscopy images available from BBBC021v1~\cite{BBBC021}, presented in Section \ref{sec_sup:biology_context}, with a batch size of 128, for 400 epoch. We use an Adam optimizer, a queue size of 1024, and set temperature at 0.07, learning rate at 0.001, and weight decay at 0.0001. Each training is made on a single V100 GPU. We perform a Kmeans (K=2) on the resulting representations of the test set of the data subsets of Nocodazole, Cytochalasin B and Taxol, and evaluate the quality of the achieved clusters compared to the ground truth using the AMI score~\cite{ami}. 

\begin{figure*}[tb]
\vskip 0.2in
\begin{center}
\centerline{\includegraphics[width=\textwidth]{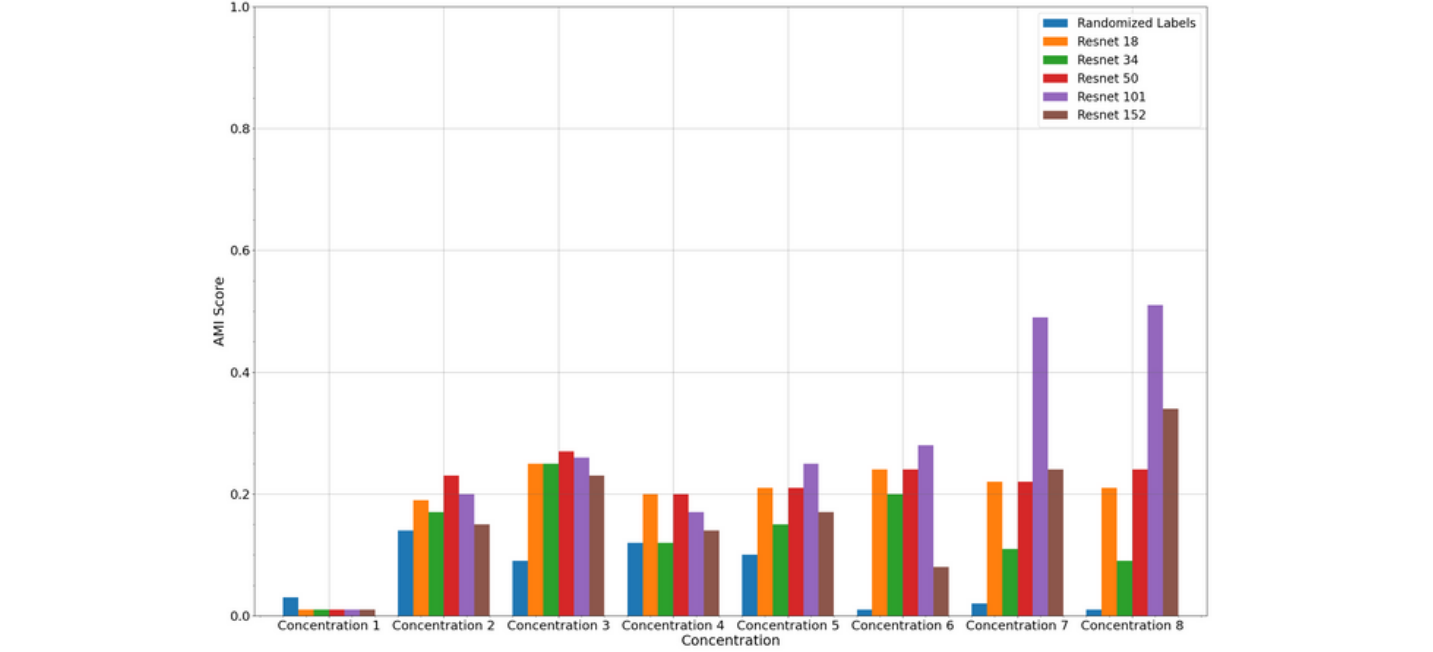}}
\caption{We perform a Kmeans clustering (K=2) on the Nocodazole and untreated images using Resnet models of various sizes, and evaluate them using the AMI score~\cite{ami} on a similar number of randomly sampled images of each concentration of Nocodazole and a similar number of untreated cell images. We observe that Resnet101 outperforms the other Resnet model sizes. We also perform an experiment to better interpret the AMI scores achieved, by randomizing the labels and performing a clustering with Resnet101, and reporting the AMI score of the clustering compared to the randomized labels.}
\label{fig:resnet}
\end{center}
\vskip -0.2in
\end{figure*}

In Table \ref{tab:benchmark_phenostyle}, we achieve the first AMI result through usage of an affine transformation composed of a rotation with an angle of 20°, a translation of 0.1 and a shear with a 10° angle, coupled with color jitter with a brightness, contrast and saturation intensity of 0.4, and a hue intensity of 0.125, with a 100\% probability, as well as random cropping of the image with a scale in the range of [0.9;1.1] and resizing to original image size of 196x196 pixel. For the second row result, we use an affine transformation composed of a rotation with an angle of 20°, a translation of 0.1 and a shear with a 10° angle, coupled with color jitter with a brightness, contrast and saturation intensity of 0.4, and a hue intensity of 0.125, with a 100\% probability, and a random rotation with a maximal angle of 360°. The results achieved through a Resnet101, are achieved by performing a Kmeans (K=2) on the representations achieved on the compound subsets using the Resnet101, and evaluating the cluster assignment quality compared to ground truth using the AMI score~\cite{ami}. The choice of Resnet101 over other Resnet sizes is motivated through testing the performance of different sizes of Resnets on the different concentrations of Nocadozole/untreated cells, on which Resnet101 consistently shows the highest performance on the 4 highest concentrations, as shown in Figure \ref{fig:resnet}.

For the clusterings in Figure \ref{fig:clustering_phenotype}, we train the same architecture with the same hyperparameters on different compositions of transformations, and perform a Kmeans (K=4) on the resulting representations of the test set. For all the clusters, the images displayed are the images closest to the centroïd of each cluster using an euclidean distance. The clustering in Figure \ref{fig:clustering_phenotype} \textit{left} is achieved by using a composition of color jitter with a brightness, contrast and saturation intensity of 0.4, and a hue intensity of 0.125, with a 100\% probability, and horizontal and vertical flips, each with 50\% probability of application, as well as random rotations with a maximal angle of 360°, an affine transformation composed of a rotation with an angle of 20°, a translation of 0.1 and a shear with a 10° angle, and a random crop with a scale sampled in the range [0.9;1.1], followed by a resizing of the image to the original size. The clustering in Figure \ref{fig:clustering_phenotype} \textit{right} is achieved by color jitter with a brightness, contrast and saturation intensity of 0.4, and a hue intensity of 0.125, with a 100\% probability, horizontal and vertical flips, each with 50\% probability of application, random rotations with a maximal angle of 360°, and a center crop with a scale of 0.5, followed by a resizing of the image to the original size. For the clustering in Figure \ref{fig:correct_phenotypes}, we trained the model with a sum of the losses (and corresponding transformations) of the models used in Figure \ref{fig:clustering_phenotype}. Through a gridsearch hyperparameter optimization, with the goal of optimizing the AMI score of a Kmeans clustering (K=2), we attributed a coefficient of 0.4 to the loss of the model used in Figure \ref{fig:clustering_phenotype} \textit{left}, and a coefficient of 1.0 to the loss of the model used in Figure \ref{fig:clustering_phenotype} \textit{right}.

\section{Additional inter-class bias results}
\label{sec_sup:additional_results}

\begin{table}[th]
\caption{The mean and standard deviation of the top1 linear evaluation accuracy, obtained through the training of a Resnet18 architecture using 12 self-supervised approaches on the Cifar10 dataset, are presented. The approaches include VicReg~\cite{vicreg}, DeepCluster v2~\cite{deepclusterv2}, SWAV~\cite{swav}, SimCLR~\cite{simclr}, SimSiam~\cite{simsiam}, MoCo~\cite{moco}, NNCLR~\cite{nnclr}, BYOL~\cite{byol}, VIBCReg~\cite{vibcreg}, Barlow Twins~\cite{barlow_twins}, and ResSL~\cite{ressl}. The experiment involves the uniform sampling of 20 values for the hue transformation probability, while maintaining a fixed maximal intensity of 0.5, and all other transformation parameters are kept constant. The results indicate that despite the variation in the transformation probability, the overall accuracy of each method remains relatively consistent, with a minimal standard deviation value.}
\label{tab:model_stats_cifar_0_1}
\vskip 0.15in
\begin{center}
\begin{small}
\resizebox{\textwidth}{!}{%
\begin{tabular}{l|c|c|c|c|c|c|c|c|c|c|c|c|}
\cline{2-13}
\multicolumn{1}{c|}{}                             & Barlow Twins                 & Byol                                 & Deep Cluster v2             & MoCo V2+                              & nnByol                      & nnclr                        & Ressl                        & SimCLR                       & SimSiam                     & SwaV                         & Vibcreg                      & Vicreg                       \\ \hline
\multicolumn{1}{|l|}{{\color[HTML]{000000} Mean}} & {\color[HTML]{343434} 89.59} & {\color[HTML]{343434} 92.09}         & {\color[HTML]{343434} 86.9} & {\color[HTML]{343434} \textbf{92.37}} & {\color[HTML]{343434} 91.3} & {\color[HTML]{343434} 89.76} & {\color[HTML]{343434} 90.21} & {\color[HTML]{343434} 90.16} & {\color[HTML]{343434} 89.6} & {\color[HTML]{343434} 86.96} & {\color[HTML]{343434} 82.47} & {\color[HTML]{343434} 89.82} \\ \hline
\multicolumn{1}{|l|}{{\color[HTML]{000000} Std}}  & {\color[HTML]{343434} 0.73}  & {\color[HTML]{343434} \textbf{0.37}} & {\color[HTML]{343434} 1.9}  & {\color[HTML]{343434} 0.44}           & {\color[HTML]{343434} 0.57} & {\color[HTML]{343434} 0.74}  & {\color[HTML]{343434} 0.85}  & {\color[HTML]{343434} 0.87}  & {\color[HTML]{343434} 1.01} & {\color[HTML]{343434} 1.2}   & {\color[HTML]{343434} 0.89}  & {\color[HTML]{343434} 0.94}  \\ \hline
\end{tabular}%
}
\end{small}
\end{center}
\vskip -0.1in
\end{table}




\end{document}